\documentclass[11pt,a4paper]{article}
\usepackage[hyperref]{emnlp2020}
\usepackage{times}
\usepackage{latexsym}
\usepackage{mwe}

\usepackage{microtype}

\aclfinalcopy 
\def\aclpaperid{510} 

\usepackage{algorithm}
\usepackage{algpseudocode}
\usepackage{amsmath}
\usepackage{booktabs}
\usepackage{comment}
\usepackage{microtype}
\usepackage{subcaption}
\usepackage{url}

\usepackage{pgfplots}
\usepgfplotslibrary{fillbetween}

\usepackage{tikz}
\usetikzlibrary{positioning, decorations.pathmorphing, shapes, arrows}

\pgfplotsset{vasymptote/.style={
		before end axis/.append code={
			\draw[densely dashed] ({rel axis cs:0,0} -| {axis cs:#1,0})
			-- ({rel axis cs:0,1} -| {axis cs:#1,0});
		}
},
width=6.5cm
}

\title{Dynamic Data Selection for Curriculum Learning via Ability Estimation}

\author{
	John P. Lalor$^1$\thanks{~~Work performed while at UMass Amherst.}, Hong Yu$^{2,3}$ \\
	$^1$Department of IT, Analytics, and Operations, University of Notre Dame \\
	$^2$ Department of Computer Science, University of Massachusetts Lowell\\
	$^3$ College of Information and Computer Sciences, University of Massachusetts Amherst\\	
	\tt{john.lalor@nd.edu}, \tt{hongyu@cs.umass.edu}
}

\date{}

\newcommand{\modelname}{Dynamic Data selection for Curriculum Learning via Ability Estimation}
\newcommand{\modelabbr}{DDaCLAE}

\begin{document}
\maketitle 
\begin{abstract}
  Curriculum learning methods typically rely on heuristics to estimate the difficulty of training examples or the ability of the model.
  In this work, we propose replacing difficulty heuristics with learned difficulty parameters. 
  We also propose \modelname~(\modelabbr), a strategy that probes model ability at each training epoch to select the best training examples at that point.
  We show that models using learned difficulty and/or ability outperform heuristic-based curriculum learning models on the GLUE classification tasks.
\end{abstract}

\section{Introduction}
\label{sec:introduction}
Curriculum learning trains a model by using easy examples first and gradually adding more difficult examples. It can speed up learning and improve generalization in supervised learning models \cite{bengio_curriculum_2009,amiri_repeat_2017,platanios_competence-based_2019}. 
A major drawback of existing curriculum learning techniques is that they rely on heuristics to measure the difficulty of data, and either ignore the competency of the model during training or rely on heuristics there as well.
For example, sentence length is often used as a proxy for difficulty in NLP tasks \cite{bengio_curriculum_2009,platanios_competence-based_2019}.
Such heuristics can be useful but have limitations.
First, the heuristic chosen may not actually be a proxy for difficulty.
Depending on the task, long sequences could signal easier or harder examples, or have no signal for difficulty.
Second, a model's notion of difficulty may not align with the heuristic imposed by a human developing the model.
It may be that examples that appear difficult for the human are in fact easy for the model to learn.

Competency was recently introduced as a mechanism to determine when new examples should be added to the training data \cite{platanios_competence-based_2019}.
However, in that work competency is a monotonically increasing function of a pre-determined initial value.
Once set, competency is not evaluated during training.
Ideally, model competency should be measured at each training epoch, so that the training data could be appropriately matched with the model at a given point in the training.
If a model is improving, then more difficult training data can be added at the next epoch. 
But if performance declines, then those difficult examples can be removed, and a smaller, easier training set can be used in the next epoch.

In this study, we propose to estimate both the difficulty of examples and the ability of deep learning models as latent variables based on model performance using Item Response Theory (IRT), a well-studied methodology in psychometrics for test set construction and subject evaluation \cite{baker_item_2004}.
IRT models estimate latent parameters such as difficulty for examples 
and a latent ability parameter for individuals (``subjects'').
IRT models are learned by administering a test to a large number of subjects, collecting and grading their responses, and using the subject-response matrix to estimate the latent traits of the data.
These learned parameters can be used to estimate the ability of future subjects, based on their graded responses to the examples.

IRT has not seen wide adoption in the machine learning community, primarily due to the fact that fitting IRT models requires a large amount of human annotated data for each example.
However, recent work has shown that IRT models can be fit using machine-generated data instead of human-generated data as input \cite{lalor_learning_2019}.

Because IRT learns example difficulty and subject ability together, 
in this work we propose replacing heuristics for learned parameters in curriculum learning.
First, we experiment with replacing a typical difficulty heuristic (sentence length) with learned difficulty parameters.
Second, we propose \modelname~(\modelabbr), a novel curriculum learning framework that uses the estimated ability of a model during the training process to dynamically identify appropriate training data.
At each training epoch, the latent ability of the model is estimated using output labels generated at the current epoch.
Once ability is known, only training data that the model has a reasonable chance of labeling correctly is included in training.
As the model improves, the estimated ability will improve, and more training examples will be added.

To the best of our knowledge, this is the first work to learn a model competency during training that is directly comparable to the difficulty of the examples.
Our study will test the following three hypotheses: \textbf{H1:} Using learned latent difficulties instead of difficulty heuristics leads to better held-out test set performance for models trained using curriculum learning, \textbf{H2:} A dynamic data selection curriculum learning strategy that probes model ability during training leads to better held-out test set performance than a static curriculum learning strategy that does not take model ability into account, \textbf{H3:} Dynamic curriculum learning is more efficient than static curriculum learning, leading to faster convergence.
We test our hypotheses on the GLUE classification data sets \cite{wang2019glue}.

Our contributions are as follows: (i) we demonstrate that for curriculum learning, learned difficulty outperforms traditional difficulty heuristics, 
(ii) we introduce a novel curriculum learning framework which automatically selects training data based on the estimated ability of the model, and (iii) we show that training using \modelabbr~leads to better performance than both traditional curriculum learning methods and a fully supervised competitive baseline. 
Our findings support the overall curriculum learning framework, and demonstrate that learning difficulty and ability lead to further performance improvements beyond common heuristics.\footnote{Code implementing our experiments and learned difficulty parameters for the GLUE data sets are available at \url{https://jplalor.github.io/irt}.}

\section{Methods}

\subsection{Curriculum Learning}

\tikzstyle{block} = [rectangle, draw, fill=blue!20, 
text centered, rounded corners]
\tikzstyle{line} = [draw, -latex']
\tikzstyle{cloud} = [draw, ellipse,fill=red!20]

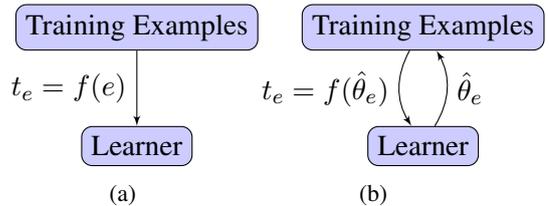
\begin{figure}
	\centering 
	\begin{subfigure}[b]{0.2\textwidth}
	\centering
    	\begin{tikzpicture}
	    \node [block] (bank) {Training Examples};
	    \node [block, below= of bank] (model) {Learner};
	    \path [line] (bank) edge [left] node {$t_e=f(e)$} (model);
	    \end{tikzpicture}%
	\caption{\label{fig:1a}}
	\end{subfigure}
	\begin{subfigure}[b]{0.2\textwidth}
	\centering 
    	\begin{tikzpicture}
	    \node [block] (bank) {Training Examples};
	    \node [block, below= of bank] (model) {Learner};
	    \path [line, bend right] (bank) edge [left] node {$t_e = f(\hat{\theta}_e)$} (model);
	    \path [line, bend right] (model) edge [right] node {$\hat{\theta}_e$} (bank);
	    \end{tikzpicture}
	\caption{\label{fig:1b}} 
	\end{subfigure}
    \caption{(\ref{fig:1a}) Traditional curriculum learning, where examples are added at each epoch according to a static monotonically-increasing schedule ($t_e = f(e)$). (\ref{fig:1b}) \modelabbr~estimates ability at each epoch ($\hat{\theta_e}$) to dynamically select appropriate training data ($t_e = f(\hat{\theta_e})$).}
	\label{fig:cl_framework} 
\end{figure}

In a traditional curriculum learning framework, training data examples are ordered according to some notion of difficulty, and the training set shown to the learner is augmented at a set pace with more and more difficult examples over time (Fig. \ref{fig:1a}).

Typically, the model's current performance is not taken into account. 
Recent work has incorporated a notion of competency to curriculum learning \cite{platanios_competence-based_2019}.
In that work the authors structure the rate at which training examples are added based on an assumption that model competency is modeled by either a linear or root function of the training epoch.\footnote{The prior work proposed other functions as well, but found that the linear and root functions performed best.}
However, there are two issues with such an approach.
First, this notion of competency is artificially rigid.
If a model's competency improves quickly, data cannot be added more quickly because the rate is predetermined.
On the other hand, if a model is slow to improve, it may struggle because data is being added too quickly.
Second, the formulation of competency proposed by the authors reduces to a competency-free curriculum learning strategy with a tuneable parameter for speed inclusion.
Once this parameter is set, there is no check of model ability during training to assess competency and update training data.
In this work our goal is to replace curriculum learning heuristics with difficulty and competency parameters learned directly using IRT (Fig. \ref{fig:1b}).

\subsection{Item Response Theory}
\label{ssec:irt} 
IRT methods learn latent parameters of test set examples (called ``items'' in the IRT literature) and latent ability parameters of individual ``subjects''.
We refer to ``items'' as ``examples'' and ``subjects'' as ``models'' respectively for clarity and consistency with the curriculum learning literature.

For a model $j$ and an example $i$, the probability that $j$ labels $i$ correctly ($z_{ij}=1$) is a function of the latent parameters of $j$ and $i$.
The one-parameter logistic (1PL) model, or Rasch model, assumes that the probability of labeling an example correctly is a function of a single latent difficulty parameter of the example, $b_i$ and a latent ability parameter of the model, $\theta_j$ \cite{rasch_studies_1960,baker_item_2004}: 

\begin{equation} 
\label{eq:irt}
p(z_{ij} = 1 \vert \theta_j, b_i) = \frac{1}{1 + e^{-(\theta_j - b_i)}}
\end{equation} 

The probability that model $j$ will label item $i$ incorrectly ($z_{ij}=0$) is: 

\begin{equation}
p(z_{ij} = 0 | \theta_j, b_i) = 1 - p(z_{ij} = 1 \vert \theta_j, b_i)
\end{equation}

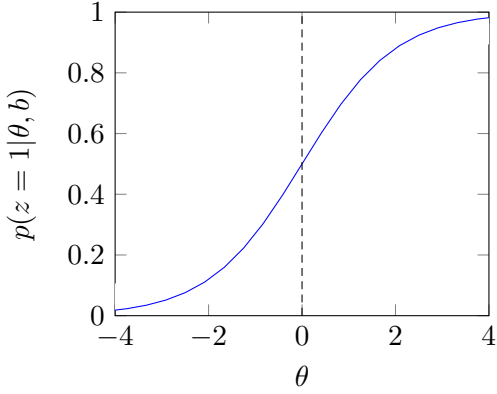
\begin{figure}
	\centering
	\begin{tikzpicture}
	\begin{axis}[xlabel=$\theta$,ylabel={$p(z=1\vert \theta,b)$}, ymin=0, ymax = 1, xmin=-4, xmax=4, vasymptote=0]
	\addplot[draw=blue, fill=gray!0] {((1.0) / (1 + exp(-1*(x ))))};
	\end{axis}
	\end{tikzpicture}
	\caption{Plot of $p(z_{ij} = 1 | \theta_j, b_i)$ as a function of $\theta$ for an example with difficulty $b=0$. Models with ability $\theta \geq 0$ (right of dashed line) have greater than 50\% chance of labeling the example correctly.}
	\label{fig:irtexamplegood}
\end{figure}

With a 1PL model, there is an intuitive relationship between difficulty and ability.
An example's difficulty value $b$ can be thought of as the point on the ability scale where a model has a 50\% chance of labeling an example correctly.
Put another way, a model has a 50\% chance of labeling an example correctly when model ability is equal to example difficulty ($\theta_j = b_i$, see Fig. \ref{fig:irtexamplegood}).

Fitting a 1PL model requires a set of $I$ examples $\{i_0, i_1, \dots, i_I\}$, a set of $J$ models $\{j_0, j_1, \dots, j_J\}$, and the binary graded responses $Z = \{\forall_{i \in I} \forall_{j \in J}: z_{ij}\}$ of the models to each of the examples.
The likelihood of a data set of response patterns $Z$ given the parameters $\Theta$ and $B$ is:
\begin{align} 
p(Z \vert \Theta, B) &= \prod_{j=1}^J \prod_{i=1}^I p(Z_{ij}=z_{ij} \vert \theta_j, b_i)
\end{align} 
where $z_{ij} = 1$ if individual $j$ answers item $i$ correctly and $z_{ij} = 0$ if they do not.


\subsection{IRT with Artificial Crowds} 
A major bottleneck of using IRT methods on machine learning data sets is the fact that each human subject would have to label all of (or most of) the examples in order to have enough response patterns to estimate the latent parameters.
Having humans annotate all of the examples in a large data set is impractical, but recent work has shown that the human subjects can be replaced with an ensemble of machine learning models \cite{lalor_learning_2019}. 
The response patterns from this ``artificial crowd'' can be used to estimate latent parameters by fitting IRT models using variational inference (VI-IRT) \cite{natesan_bayesian_2016,lalor_learning_2019}. 


VI-IRT approximates the joint posterior distribution $\pi(\Theta,B|Z)$ by the variational distribution:

\begin{align} 
q(\Theta, B) &=  \prod_{j=1}^J \pi^\theta_j(\theta_j) \prod_{i=1}^I \pi^b_i(b_i)
\end{align} 
where $\pi^\theta_j()$ and $\pi^b_i()$ denote Gaussian densities for different parameters.
Parameter means and variances are determined by minimizing the KL-Divergence between $q(\Theta,B)$ and $\pi(\Theta,B|Z)$.

In selecting priors for VI-IRT we follow the results of prior work where hierarchical priors were used \cite{natesan_bayesian_2016,lalor_learning_2019}.
The hierarchical model assumes that ability and difficulty means are sampled from a vague Gaussian prior, and ability and difficulty variances are sampled from an inverse Gamma distribution:
\begin{align*}
\theta_j\ |\ m_\theta, u_\theta &\sim N(m_{\theta}, u^{-1}_{\theta}) \\
b_i\ |\ m_b, u_b &\sim N(m_b, u^{-1}_b) \\
m_{\theta}, m_{b} &\sim N(0, 10^6) \\
u_{\theta}, u_b &\sim \Gamma(1, 1)
\end{align*}

\subsubsection{Estimating Model Ability}
\label{ssec:scoring}
Estimating the ability of a model at a point in time is done with a ``scoring'' function. 
When example difficulties are known, model ability is estimated by maximizing the likelihood of the data given the response patterns and the example difficulties to obtain the ability estimate.
All that is required is a single forward pass of the model on the data, as is typically done with a test or validation set. 
\begin{align}
Z_j &= \forall_{y \in Y} \mathbf{I}[y_i = \hat{y_i}] \\
\hat{\theta}_j &= \operatorname*{arg\,max}_{\theta_j}  \prod_{i=1}^I p(z_{ij}=y_{ij} \vert \theta_j) 
\end{align}

\subsection{\modelname}
\label{ssec:dcl}

We propose \modelabbr, where training examples are selected dynamically at each training epoch based on the estimated ability of the model at that epoch.
With \modelabbr, model ability can be estimated according to a well-studied psychometric framework as opposed to heuristics.
The estimated ability of the model at a given epoch $e$, $\hat{\theta}_e$, is on the same scale as the difficulty parameters of the data, so there is a principled approach for selecting data at any given training epoch.

Algorithm \ref{alg:dcl} describes the training procedure.
The first step of \modelabbr~is to estimate the ability of the model using the scoring function (\S \ref{ssec:scoring}, Alg. \ref{alg:dcl} line 2). 
To do this we use the full training set, but crucially, only to get response data, not to update parameters (i.e., no backward pass). 
We do not use a held out development set for estimating ability because we do not want the development set to influence training.
In our experiments the development set is only used for early stopping.
Model outputs are obtained for the training set, and graded as correct or incorrect as compared to the gold standard label (Alg. \ref{alg:dcl} line 8). 
This response pattern is then used to estimate model ability at the current epoch ($\hat{\theta}_e$, Alg. \ref{alg:dcl} line 9).

Once ability is estimated, data selection is done by comparing estimated ability to the examples' difficulty parameters.
Each example in the training examples has an estimated difficulty parameter ($b_x$).
If the difficulty of an example is less than or equal to the estimated ability, then the example is included in training for this epoch.
Examples where the difficulty is greater than estimated ability are not included (Alg. \ref{alg:dcl} line4).
Finally, the model is trained with the training data subset (Alg. \ref{alg:dcl} line 5).

With \modelabbr, the training data size does not have to be monotonically increasing. 
\modelabbr~adds or removes training data based not on a fixed step schedule but rather by probing the model at each epoch and using the estimated ability to match data to the model (Figure~\ref{fig:cl_framework}).
This way if a model has a high estimated ability early in training, then more data can be added to the training set more quickly, and learning isn't artificially slowed down due to the curriculum schedule.
If a model's performance suffers when adding data too quickly, then this will be reflected in lower ability estimates, which leads to less data selected in the next epoch. 

\begin{algorithm}[t!]
	\small 
	\caption{\modelabbr}
	\hspace*{\algorithmicindent}\textbf{Input:} Data (X, Y), model $\phi$, difficulties D, $\text{num\_epochs}$ \\
	\hspace*{\algorithmicindent}\textbf{Output:} Learned model $\phi$ 
	\begin{algorithmic}[1]
		\For{$e$ in $\text{num\_epochs}$}
		\State $\hat{Y} = \phi(X)$
		\State $\hat{\theta}_e = score(Y, \hat{Y}, D)$
		\State $X_e, Y_e = \{(x,y): b_x \leq \hat{\theta}_e\}$
		\State $train(\phi, X_e, Y_e)$
		\EndFor 
		\Procedure{score}{$Y, \hat{Y}, D$}
		\State $Z = \forall_{y \in Y} \mathbf{I}[y_i = \hat{y_i}]$
		\State $\hat{\theta}_e = \operatorname*{arg\,max}_\theta p(Z \vert \theta, b)$
		\State return $\hat{\theta_e}$
		\EndProcedure 
	\end{algorithmic} 
	\label{alg:dcl}
\end{algorithm} 

\section{Data and Experiments} 


\subsection{Data}
\label{ssec:data_description}
For our experiments we consider the GLUE English-language classification tasks \cite{wang2019glue}: MNLI \cite{williams2018broad}, QQP,\footnote{\url{https://data.quora.com/First-Quora-Dataset-Release-Question-Pairs}} QNLI \cite{rajpurkar2016squad,wang2019glue}, SST-2 \cite{socher_recursive_2013}, MRPC \cite{dolan2005automatically}, and RTE \cite{bentivogli2009fifth}.
Data set summary statistics are provided at Table \ref{tab:glue_stats}.
We exclude the WNLI data set.\footnote{See \url{https://gluebenchmark.com/faq} note 12.}

Because test set labels for our tasks are only available via the GLUE evaluation server, we use the held-out development sets to measure performance.
For training, we do a 90\%-10\% split of the training data, and use the 10\% split as our held out development set for early stopping.
We can then use the full development set as our test set to evaluate performance without making multiple submissions to the GLUE server.

\begin{table}[h!]
	\centering
	\footnotesize 
	\begin{tabular}{lccc}
		\toprule
		Data set & Train & Dev & Test \\
		\midrule
		MNLI &353k&39k&9.8k \\
		MRPC &3.3k&366&409 \\
		QNLI &94k&10k&5.5k \\
		QQP &327k&36k&40k \\
		RTE &2.2k&249&278 \\
		SST-2 &61k&6.7k&873 \\
		\bottomrule
	\end{tabular}
	\caption{Statistics for our experiments. Note that these values differ from the GLUE server data (see \S \ref{ssec:data_description}).} 
	\label{tab:glue_stats}
\end{table}

\subsection{Generating Response Patterns}

To learn the difficulty parameters of the data we require a data set of response patterns.
Gathering enough labels for each example to fit an IRT model would be prohibitively expensive for human annotators.
In addition, the annotation quality may be suspect due to the humans labeling tens of thousands of examples.
Therefore we used artificial crowds to generate our response patterns. 
Prior work has shown that this is an effective way to generate a set of response patterns for fitting IRT models to machine learning data \cite{lalor_learning_2019}.

Briefly, for each data set an ensemble of neural network models are trained with different subsets of the training data.
Training data is subsampled and corrupted via label flipping so that performance across models in the ensemble is varied.
Each trained model then labels all of the examples (train/validation/test), which are graded against the gold-standard label.
The output response patterns are used to fit an IRT model for the data (\S \ref{ssec:irt}).

\subsection{Experiments} 
\label{ssec:experiments}

To demonstrate the effectiveness of \modelabbr~we compare against a fully supervised baseline as well as a competence-based method (CB) that uses a fixed, monotonically-increasing schedule to add examples during training~\cite{platanios_competence-based_2019}.
All experiments described were run five times.
Average performance and 95\% CI are reported below.

For each task, we trained two standard model architecture for a set number of epochs: $\text{BERT}_{base}$ and LSTM. 
We use the $\text{BERT}_{\text{base}}$ model \cite{devlin_bert:_2018} as implemented by HuggingFace.\footnote{\url{https://github.com/huggingface/pytorch-transformers}}
Each model was trained for 10 epochs, with a learning rate of 2e-5 and a batch size of 8. 
Dropout for all fine-tuning layers was set to $0.1$.
We used gradient norm clipping at $1$ to avoid exploding gradients.

The LSTM model consists of a 300D LSTM sequence-embedding layer \cite{hochreiter_long_1997} (one or two LSTMs for single- and two-sentence tasks, respectively).
The sentence encodings are then concatenated and passed through three tanh layers.
Finally, the output is passed to a softmax classifier layer to output class probabilities.
The models were implemented in DyNet \cite{dynet}.
Models were trained with SGD for 100 epochs with a learning rate of 0.1, and dev set accuracy was used for early stopping.

Training data available to the model at each epoch varied according to the curriculum applied:

\begin{enumerate}
	\item 
	Fully Supervised: At each epoch, the model has access to the full training set
	\item 
	CB-Linear: The proportion of training examples to include at time $t$ is 
	
	$c_{linear}(t) \overset{\Delta}{=} \text{min} (1, t\frac{1-c_0}{T} + c_0)$
	
	\item 
	CB-Root: The proportion of training examples to include at time $t$ is 
	
	$c_{sqrt}(t) \overset{\Delta}{=} \text{min}(1, \sqrt{t\frac{1-c_0^2}{T} + c_0^2})$

	\item 
	\modelabbr: At each epoch, model ability is estimated ($\hat{\theta}_e$, see \S \ref{ssec:dcl}) and all examples where $b_x \leq \hat{\theta}_e$ are included
\end{enumerate}

\begin{table*}[ht]
\centering
\footnotesize
\begin{tabular}{lcccccc}
  \toprule
 & MNLI & MRPC & QNLI & QQP & RTE & SST2 \\ 
  \midrule
Fully Supervised & 72.8 [$\pm$13.1] & 76.76 [$\pm$4.19] & 88.32 [$\pm$1.86] & 84.87 [$\pm$7.08] & 59.35 [$\pm$3.55] & 87.55 [$\pm$4.77] \\ 
  CB Lin. ($d_{length}$) & 72.37 [$\pm$12.9] & 74.95 [$\pm$3.65] & 81.63 [$\pm$10.5] & 90.07 [$\pm$0.22] & 58.77 [$\pm$2.55] & 73.78 [$\pm$12.2] \\ 
  CB Lin. ($d_{irt}$) $\clubsuit$ & 82.36 [$\pm$0.28] & 84.85 [$\pm$1.24] & 87.9 [$\pm$1.14] & 90.13 [$\pm$0.06] & 58.27 [$\pm$3.46] & 90.05 [$\pm$0.26] \\   
  CB Root ($d_{length}$) & 82.17 [$\pm$0.17] & 77.11 [$\pm$3.37] & 81.74 [$\pm$10.6] & 84.79 [$\pm$6.97] & 57.26 [$\pm$2.82] & 76.93 [$\pm$11] \\ 
  CB Root ($d_{irt}$) $\clubsuit$& \bf \underline{82.46} [$\pm$0.12] & \bf \underline{84.95} [$\pm$0.96] & \underline{88.48} [$\pm$1.16] & \bf \underline{90.27} [$\pm$0.05] & \bf \underline{62.67} [$\pm$1.4] & \underline{90.73} [$\pm$0.22] \\ 
  DDaCLAE $\clubsuit$ & 81.65 [$\pm$0.37] & 81.37 [$\pm$2.19] & \bf 89.33 [$\pm$0.75] & 90.15 [$\pm$0.13] & 59.71 [$\pm$2.63] & \bf 90.99 [$\pm$0.37] \\ 
   \bottomrule
\end{tabular}
\caption{BERT dev set accuracy results, including 95\% CI, for each task under consideration. During training, 10\% of the training set was held out and used for early stopping. Highest overall accuracy is \textbf{bolded}. Highest accuracy among competence-based methods is \underline{underlined}. $\clubsuit$ indicates newly proposed methods.} 
\label{tab:acc_bert-True}
\end{table*}

\begin{table*}[ht]
\centering
\footnotesize
\begin{tabular}{lcccccc}
  \toprule
 & MNLI & MRPC & QNLI & QQP & RTE & SST2 \\ 
  \midrule
Fully Supervised & 64.96 [$\pm$0.18] & 69.04 [$\pm$0.71] & 62.72 [$\pm$0.48] & 81 [$\pm$0.12] & 50.12 [$\pm$0.68] & 84.73 [$\pm$0.34] \\ 
  CB Lin. ($d_{length}$) & 63.79 [$\pm$0.23] & 68.38 [$\pm$0.55] & 58.97 [$\pm$0.84] & 79.99 [$\pm$0.21] & 49.96 [$\pm$1.2] & 83.97 [$\pm$0.39] \\ 
  CB Lin. ($d_{irt}$) $\clubsuit$& 65.18 [$\pm$0.14] & \underline{70.18} [$\pm$0.5] & 61.21 [$\pm$0.5] & 81.68 [$\pm$0.11] & 50.45 [$\pm$1.06] & 84.71 [$\pm$0.35] \\ 
  CB Root ($d_{length}$) & 64.05 [$\pm$0.39] & 69.36 [$\pm$0.35] & 60.04 [$\pm$0.73] & 79.23 [$\pm$0.55] & 50.32 [$\pm$1.62] & 83.74 [$\pm$0.43] \\ 
  CB Root ($d_{irt}$) $\clubsuit$& \underline{65.37} [$\pm$0.16] & 69.98 [$\pm$0.52] & \underline{61.87} [$\pm$0.27] & \bf \underline{81.91} [$\pm$0.08] & \underline{50.93} [$\pm$0.69] & \underline{84.74} [$\pm$0.12] \\ 
  DDaCLAE $\clubsuit$& \bf 65.93 [$\pm$0.24] & \bf 70.38 [$\pm$0.62] & \bf 63.01 [$\pm$0.4] & 81.75 [$\pm$0.11] & \bf 52.35 [$\pm$1.2] & \bf 85.7 [$\pm$0.19] \\ 
   \bottomrule
\end{tabular}
\caption{LSTM dev set accuracy results, including 95\% CI, for each task under consideration. During training, 10\% of the training set was held out and used for early stopping. Highest overall accuracy is \textbf{bolded}. Highest accuracy among competence-based methods is \underline{underlined}. $\clubsuit$ indicates newly proposed methods.} 
\label{tab:acc_lstm-True}
\end{table*}

For the competence-based methods, $t$ is the current time-step in training, $T$ is the point where the model is fully competent, $c_0$ is the initial competency. 
We set $c_0=0.01$ as per the original paper and set $T$ to be equal to $\frac{numepochs}{2}$ \cite{platanios_competence-based_2019}.
The competence-based models reach ``competency'' halfway through training and train with the full training set for the second half.

To determine the effectiveness of difficulty as estimated by IRT methods, we experiment with two versions of the competency-based models in our NLP tasks:
(1) $d_{length}$: using sentence length as a heuristic for difficulty, as in the prior work \cite{platanios_competence-based_2019}. 
For sentence-pair tasks such as MNLI we use the length of the first sentence for $d_{length}$.
(2) $d_{irt}$: difficulty as estimated by fitting an IRT model using the artificial crowd (\S \ref{ssec:irt}). 
$d_{length}$ is one of two common heuristics used for difficulty in prior work.
Word rarity, where the negative log likelihood of the sentence is computed, has also been proposed as a heuristic \cite{platanios_competence-based_2019}.
Recent empirical evaluations have shown that word rarity and sentence length perform similarly as heuristics, and we therefore use sentence length as our heuristic for comparison \cite{platanios_competence-based_2019}.

It is worth noting here that neither CB-Linear nor CB-Root actually measure competency of the model at any point. 
Instead it is assumed that the model becomes more and more competent over time, whereas with \modelabbr~model competency is probed at each training epoch and training data is selected based on this competency.

\paragraph{Estimating Ability}
For \modelabbr, there is a potentially significant cost associated with estimating $\theta_e$. 
Estimating $\theta_e$ requires an additional forward pass through the training data set to gather the labels for scoring as well as MLE estimation (\ref{ssec:scoring}).
For large data sets this can effectively double the number of forward passes during training.
To alleviate the extra cost, we sample from the training set before our first epoch, and use this down-sampled subset as our ability estimation set.
As most examples have difficulty values between $-3$ and $3$, the full training set isn't necessary for estimating $\theta_e$.
For our experiments we sampled $1000$ examples for ability estimation, significantly reducing the cost.
Identifying the optimal number of examples needed to estimate ability is left for future work.

\section{Results}

\subsection{Analysis of Difficulty Heuristic}

To explore the discrepancy between common difficulty heuristics and a learned difficulty parameter, we calculated the Spearman rank-order correlation between difficulty using the sentence-length heuristic, $d_{length}$, and difficulty as estimated by IRT, $d_{irt}$ (Table \ref{tab:correlations}).
In most cases, correlation is around $0$, indicating no (or minimal) correlation between the two values. 
In fact for some tasks the correlation is negative (e.g., $-0.19$ for MRPC).
For SST-2, there is a moderate positive correlation, indicating that some short examples are easier for the task of sentiment analysis.
However, the lack of (or negative) correlation in other tasks indicates that sentence length, a common heuristic for difficulty in curriculum learning work, is not aligned with the more theoretically-grounded difficulty estimates of IRT.

\begin{table}[h!]
	\centering 
	\small 
	\begin{tabular}{cccccccc}
		\toprule
		  & SST-2 & MRPC & QNLI & QQP & RTE & MNLI\\
		\midrule
		$\rho$ & 0.21 & -0.19 & 0.01 & -0.09 & -0.02 & 0.03\\ 
		\bottomrule 
	\end{tabular}
	\caption{Spearman rank-order correlations between length difficulty heuristic and IRT-estimated difficulty.}
	\label{tab:correlations}
\end{table}

\subsection{Replacing Difficulty Heuristics (H1)}
By replacing difficulty heuristics with learned difficulty parameters in a static curriculum learning framework, we see that performance is improved for all GLUE classification tasks, with both BERT and LSTM models (Tables \ref{tab:acc_bert-True} and \ref{tab:acc_lstm-True}).
Using learned difficulty parameters outperforms both the fully supervised baseline and the equivalent curriculum learning strategy with difficulty heuristics (e.g., comparing CB Root ($d_{irt}$) with CB Root ($d_{length}$)).

This result confirms our first hypothesis (\textbf{H1}) and demonstrates that learning difficulty parameters for data leads to more effective curriculum learning models.
The models are able to leverage a theoretically-based difficulty metric instead of a heuristic such as sentence length.

\begin{table*}[t!]
\centering
\begingroup\small
\begin{tabular}{lcccccc}
  \toprule
 & MNLI & MRPC & QNLI & QQP & RTE & SST2 \\ 
  \midrule
Fully Supervised & \bf 1.8 [$\pm$0.26] & \bf 5.2 [$\pm$1.51] & 3.4 [$\pm$0.78] & \bf 3.6 [$\pm$1.14] & \bf 3.2 [$\pm$1.12] & \bf 2.6 [$\pm$0.78] \\ 
  CB Lin. ($d_{length}$) & \underline{6} [$\pm$0.41] & \underline{6.6} [$\pm$1.78] & 5.4 [$\pm$1.47] & 7.8 [$\pm$0.49] & 8.4 [$\pm$0.67] & 5.6 [$\pm$1.73] \\ 
  CB Lin. ($d_{irt}$) $\clubsuit$ & \underline{6} [$\pm$0] & 7.6 [$\pm$1.28] & 5.6 [$\pm$0.32] & 8 [$\pm$0.41] & 6.2 [$\pm$1.51] & 8 [$\pm$0.83] \\ 
  CB Root ($d_{length}$) & \underline{6} [$\pm$0.72] & 7 [$\pm$1.8] & 4.4 [$\pm$1.14] & 7.4 [$\pm$1.28] & \underline{5} [$\pm$1.89] & \underline{5.2} [$\pm$1.57] \\ 
  CB Root ($d_{irt}$) $\clubsuit$& 6.2 [$\pm$0.26] & 7 [$\pm$1.17] & 6 [$\pm$0.72] & 7.8 [$\pm$0.64] & 8 [$\pm$0.72] & 7.8 [$\pm$0.96] \\ 
  DDaCLAE $\clubsuit$& 8.4 [$\pm$0.98] & 9.4 [$\pm$0.32] & \bf \underline{2} [$\pm$0.41] & \underline{5.8} [$\pm$0.76] & 6.2 [$\pm$1.57] & 5.8 [$\pm$0.76] \\ 
   \bottomrule
\end{tabular}
\endgroup
\caption{Mean number of training epochs to convergence (lower is better) for BERT models, with 95\% CI. The best overall is \textbf{bolded}, and the best among the CB methods is \underline{underlined}. $\clubsuit$ indicates newly proposed methods.} 
\label{tab:epoch_bert-True}
\end{table*}

\begin{table*}[t!]
\centering
\footnotesize
\begin{tabular}{lcccccc}
  \toprule
 & MNLI & MRPC & QNLI & QQP & RTE & SST2 \\ 
  \midrule
  Fully Supervised & \bf 17.25 [$\pm$1.7] & \bf 14.17 [$\pm$4.13] & \bf 19.17 [$\pm$5.03] & 78.6 [$\pm$15.2] & 58 [$\pm$6.18] & \bf 43.67 [$\pm$8.68] \\ 
  CB Lin. ($d_{length}$) & 37 [$\pm$0.72] & 59.2 [$\pm$3.39] & 85.6 [$\pm$16.54] & 80 [$\pm$10.54] & 53.8 [$\pm$10.93] & 73.6 [$\pm$6.48] \\ 
  CB Lin. ($d_{irt}$) $\clubsuit$ & 39 [$\pm$0.96] & 37.33 [$\pm$3.89] & 55.08 [$\pm$8.15] & 86.1 [$\pm$5.25] & 63.92 [$\pm$9.44] & 77.33 [$\pm$4.44] \\ 
  CB Root ($d_{length}$) & 30.67 [$\pm$1.1] & 39.4 [$\pm$1.41] & 69.2 [$\pm$21.02] & 77.5 [$\pm$13.1] & 43.6 [$\pm$11.41] & 63 [$\pm$1.09] \\ 
  CB Root ($d_{irt}$) $\clubsuit$ & 29.5 [$\pm$0.8] & \underline{30.5} [$\pm$4.93] & 37.75 [$\pm$6.86] & 80.7 [$\pm$7.11] & 56.17 [$\pm$10.19] & 68.42 [$\pm$3.85] \\ 
  DDaCLAE $\clubsuit$ & \underline{20.2} [$\pm$1.33] & 46.33 [$\pm$13.1] & \underline{23.8} [$\pm$2] & \bf \underline{68.6} [$\pm$14.8] & \bf \underline{26.17} [$\pm$14.29] & \underline{50.33} [$\pm$6.68] \\ 
   \bottomrule
\end{tabular}
\caption{Mean number of training epochs to convergence (lower is better) for LSTM models, with 95\% CI. The best overall is \textbf{bolded}, and the best among the CB methods is \underline{underlined}. $\clubsuit$ indicates newly proposed methods.} 
\label{tab:epoch_lstm-True}
\end{table*}

\subsection{Dynamic Curriculum Learning (H2)}
For the $\text{BERT}_{base}$ model, \modelabbr~outperforms the fully supervised baseline as well as all other curriculum learning methods on 2 of the 6 tasks (QNLI and SST-2, Table \ref{tab:acc_bert-True}). 
However, we find that \modelabbr~does not lead to further performance improvements on the other tasks for $\text{BERT}_{base}$.

\modelabbr~does give the best performance for 5 of the 6 GLUE tasks (all except QQP) when used to train the LSTM model (Table \ref{tab:acc_lstm-True}).
This could be due to the fact that training the $\text{BERT}_{base}$ models is a \textit{fine-tuning} procedure against the already pre-trained models.
Therefore there is not much room for performance improvement switching from a static to a dynamic curriculum learning model. 
On the other hand, the LSTM models are all randomly initialized, and therefore require a full training procedure.
In scenarios like this, \modelabbr~is an effective procedure to improve performance.
With \modelabbr, the model is trained using data that is most appropriate for its current ability.
Examples that are too hard are not included too early.

One potential issue with \modelabbr~is the chance of a high variance model, due to the additional step of estimating model ability during training.
However we find that variance in terms of output performance is quite low for both $\text{BERT}_{base}$ and LSTM models trained with \modelabbr.

\subsection{Training Efficiency (H3)}

In addition to test-set performance, we analyzed the efficiency of the curriculum learning training methods.
For each experiment, we calculated the average number of training epochs required to reach the point of early stopping (based on held out dev set accuracy).
For $\text{BERT}_{base}$, fully supervised training is almost always the most efficient (Table \ref{tab:epoch_bert-True}).
This should not be surprising, as the model is already pre-trained, and fine-tuning only requires a small number of passes over the task data.

For training the LSTM model, efficiency results are more mixed (Table \ref{tab:epoch_lstm-True}).
In most cases the fully-supervised training is again most efficient, however \modelabbr~does not incur significant efficiency costs.
For QQP and RTE, \modelabbr~is the most efficient training paradigm.
For MNLI, QNLI, and SST-2, \modelabbr~efficiency is within the 95\% CI of the baseline results.
Recall that for the LSTM model \modelabbr~is the most effective in terms of test set accuracy as well, so we can say that the improved test set performance does not come at the cost of training time efficiency.

 

\subsection{Distribution of Difficulty}

\begin{figure}[h!]
	\centering
	\begin{subfigure}[b]{0.49\columnwidth}
		\centering
		\includegraphics[width=\columnwidth]{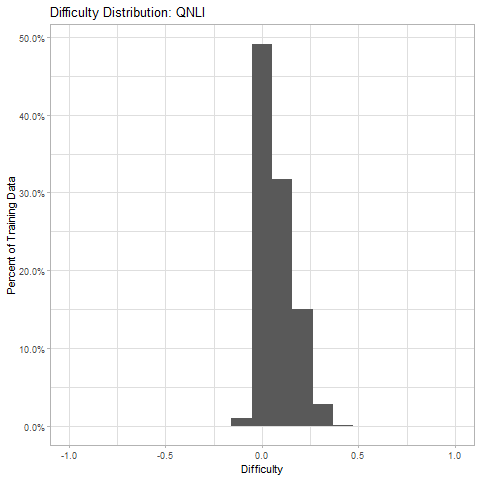}
		\caption{QNLI}\label{fig:qnli} 
	\end{subfigure} 
	\begin{subfigure}[b]{0.49\columnwidth}
		\centering
		\includegraphics[width=\columnwidth]{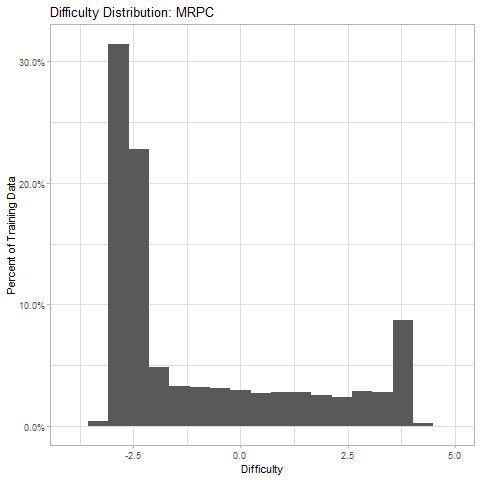}
		\caption{MRPC}\label{fig:MRPC} 
	\end{subfigure} 
	\caption{Percentage plot of difficulty values in QNLI and MRPC data sets. Please note that x-axes are intentionally scaled to reflect the range for a given data set, and are not consistent between plots. Additional plots for other tasks included in the supplemental material.}
	\label{fig:diffs}
\end{figure}

Figure \ref{fig:diffs} shows percentage plots of estimated difficulty for two of the GLUE classification tasks, QNLI and MRPC.
As the plots show, the distribution in difficulty varies between the tasks.
For MRPC, there are more difficult examples, percentage-wise, than in the QNLI data set.
This reflects the current state of the GLUE leaderboard, where the top-performing model accuracies are 97.8\% and 92.6\% on QNLI and MRPC, respectively.\footnote{Scores as of the time of this submission.}
This is also reflected in our results, where model performance is higher for QNLI than MRPC (Table \ref{tab:acc_bert-True}).
Knowing the distribution of difficulty in a data set is useful information for model development and evaluation strategies.
In the case of curriculum learning we leverage this learned difficulty to train our models.

\subsection{Additional Training Time Considerations}

Estimating model ability every training epoch with \modelabbr~can potentially increase training time significantly. 
If at each training epoch there is a need to run a full MLE optimization, the cost in terms of time could significantly outweigh performance improvements.
To mitigate this we randomly sample examples for ability estimation.

Comparing training with \modelabbr~to training a fully-supervised baseline, the average impact on training time ranges from an additional few minutes for smaller data sets (e.g., MRPC) to an additional few hours for the larger data sets (e.g., MNLI). 
This impact grows with the data set size because when estimating ability, all of the training examples are used to generate a response pattern, then a subset of $1000$ are selected for estimating ability. 
Future implementations can sample the training data before gathering response patterns, or pre-select a subset with varying difficulty parameters and to use as a static ``probe set'' to estimate ability at each epoch.

\section{Related work}

Curriculum learning is a well-studied area of machine learning \cite{bengio_curriculum_2009}.
The primary focus has been on developing new heuristics to identify easy and difficult examples in order to build a curriculum. 
Originally, curriculum learning methods were evaluated on toy data sets with heuristic measures of difficulty \cite{bengio_curriculum_2009}.
For example, on a shapes data set, shapes with more sides were considered difficult.
Similarly, longer sentences were considered difficult.
Word rarity has also been proposed as a heuristic for difficulty, with similar results to sentence length \cite{platanios_competence-based_2019}.
Recent work on automating curriculum learning strategies use multi-arm bandits to minimize regret with respect to curriculum selection \cite{graves2017automated}.
In that work the authors again rely on proxies for progress (loss-driven and complexity-driven). 
Loss-driven proxies are inherently model-specific, in that the difficulty of an example is determined by a specific model's performance on the example. 
Using a global difficulty such as one learned using IRT methods allows for an interpretable difficulty metric that applies across models.
The complexity-driven proxies proposed are specific to neural networks, while~\modelabbr is a generic algorithm for dynamic curriculum learning.

Spaced repetition strategies (SR) can be effective for improving model performance \cite{amiri_repeat_2017,amiri_neural_2019}.
Instead of using a traditional curriculum learning setup, spaced repetition bins examples based on estimated difficulty, and shows bins to the model at differing intervals so that harder examples are seen more frequently than easier ones.
This method has been shown to be effective for human learning, and results demonstrate effectiveness on NLP tasks as well.
Similarly to traditional curriculum learning, SR uses heuristics for difficulty and rigid schedulers to determine when examples should be re-introduced to the learner.

Self-paced learning (SPL) is another related strategy for example ordering during training \cite{kumar_self-paced_2010}.
In our work the difficulty values of examples are global and static, where for SPL examples do not have set difficulty values; instead, groups of examples are considered in sets.
In addition, model competency is not considered in SPL, it is assumed that competency improves as more difficult examples are added.


There has been recent work investigating the theory behind curriculum learning \cite{weinshall_curriculum_2018-1,hacohen_power_2019-2}, particularly around trying to define an ideal curriculum.
The authors explicitly identify the two key aspects of curriculum learning, namely ``sorting by difficulty'' and ``pacing.''
Curriculum learning theoretically leads to a steeper optimization landscape (i.e., faster learning) while keeping the same global minimum of the task without curriculum learning.
In that work there is still a reliance on ``pacing functions'' as opposed to an actual assessment of model ability at a point in time.
This work may open interesting new areas of theoretical study linking difficulty and ability in curriculum learning.

Theoretical results \cite{hacohen_power_2019-2} have also demonstrated a key distinction between curriculum learning and similar methods such as self-paced learning \cite{kumar_self-paced_2010}, hard example mining \cite{shrivastava_training_2016}, and boosting \cite{freund_decision-theoretic_1997}: namely that the former considers difficulty with respect to the final hypothesis space (i.e., a model trained on the full data set) while the later methods consider ranking examples according to how difficult the current model determines them to be.
\modelabbr~bridges a gap between these methods by probing model ability at the current point in training and using this ability to identify appropriate training examples in terms of global difficulty.

In addition, a key component of most prior work in curriculum learning is the notion of balance.
When defining a curriculum, it is often the case that proportions are maintained between classes.
That is, difficulty itself is not the only factor when building the curriculum.
Instead, the easiest examples for each class are added so that the model is proportionally exposed to the data consistent with the full training set.
\modelabbr~does not consider class labels when selecting examples for training.
It is important to note here that labels are used when learning difficulties, estimating ability, and actually updating parameters during training. They are not used to balance the curriculum.
In this way \modelabbr~is more closely aligned with a pure curriculum learning strategy that considers only the easiness/hardness of an example during training.
This is an added benefit to \modelabbr~as there is no need for class label accounting during training. 

\section{Conclusion} 

This work validates and supports the existing literature on curriculum learning.
Our results confirm that curriculum learning methods for supervised learning can lead to faster convergence or better local minima, as measured by test set performance \cite{bengio_curriculum_2009}.
We have shown that by replacing a heuristic for difficulty with a theoretically-based, learned difficulty value for training examples, static curriculum learning strategies can be improved.
We have also proposed \modelabbr, the first curriculum learning method to dynamically probe a model during training to estimate model ability at a point in time.
Knowing the model's ability allows for data to be selected for training that is appropriate for the model and is not rigidly tied to a heuristic schedule.
\modelabbr~trains more effective models in most cases that we considered, particularly for randomly initialized LSTM models.

Based on our experiments, we report mixed results on our stated hypotheses.
Replacing heuristics with learned difficulty values leads to improved performance when training models with curriculum learning, supporting \textbf{H1}.
\modelabbr~does outperform other training setups when used to train LSTM models.
Results are mixed when used to fine-tune the $\text{BERT}_{base}$ model.
Therefore \textbf{H2} is partially supported.
We see similarly mixed results when evaluating efficiency.
With $\text{BERT}_{base}$ fine-tuning, fully supervised fine-tuning is usually the most efficient, as the number of fine-tuning epochs needed is already very low.
For LSTM, \modelabbr~is more efficient than the other curriculum learning strategies, and is the most efficient overall for two of the six tasks.
\textbf{H3} is partially supported by these results.

Even though it is dynamic, \modelabbr~employs a simple schedule: only include examples where $b_x \leq \hat{\theta}_e$.
However, being able to estimate ability on the fly with \modelabbr~opens up an exciting new research direction: what is the best way to build a curriculum, knowing example difficulty and model ability (e.g., the 85\% rule of~\citealp{Wilson255182})?




\section*{Acknowledgements}
The authors would like to thank Hao Wu and Hadi Amiri for their helpful conversations with regards to this work.
We thank the anonymous reviewers for their comments and suggestions. 
This work was supported in part by LM012817 from the National Institutes of Health.
This work was also supported in part by the Center for Intelligent Information Retrieval at UMass Amherst, and the Center for Research Computing and Mendoza College of Business at the University of Notre Dame.
The contents of this paper do not represent the views of CIIR, NIH, VA, Notre Dame, University of Massachusetts Lowell or the United States Government.

\bibliographystyle{acl_natbib}
\bibliography{library}

\end{document}